\documentclass[conference]{IEEEtran}
\usepackage{amsmath, amssymb, amsfonts}
\usepackage{graphicx, url}
\usepackage{booktabs}
\usepackage{cite}
\usepackage{pgfplots}
\pgfplotsset{compat=1.18}
\usepackage{longtable}
\title{Statistical and Machine Learning Analysis of Traffic Accidents on US 158 in Currituck County: A Comparison with HSM Predictions}
\author{
    \IEEEauthorblockN{Jennifer Sawyer\IEEEauthorrefmark{1}, Julian Allagan\IEEEauthorrefmark{2}}
    \IEEEauthorblockA{\IEEEauthorrefmark{1}jnsawyer@ecsu.edu}
    \IEEEauthorblockA{\IEEEauthorrefmark{2}adallagan@ecsu.edu}
    \IEEEauthorblockA{\textit{Department of Mathematics, Computer Science, and Engineering Technology} \\ \textit{Elizabeth City State University} \\ North Carolina, USA}
}
\begin{document}

\maketitle

\begin{abstract}
This study extends previous hotspot and Chi-Square analysis by Sawyer \cite{sawyer2025hotspot} by integrating advanced statistical analysis, machine learning, and spatial modeling techniques to analyze five years (2019--2023) of traffic accident data from an 8.4-mile stretch of US 158 in Currituck County, NC. Building upon foundational statistical work, we apply Kernel Density Estimation (KDE), Negative Binomial Regression, Random Forest classification, and Highway Safety Manual (HSM) Safety Performance Function (SPF) comparisons to identify comprehensive temporal and spatial crash patterns. A Random Forest classifier predicts injury severity with 67\%  accuracy, outperforming HSM SPF. Spatial clustering is confirmed via Moran's I test ($I = 0.32$, $p < 0.001$), and KDE analysis reveals hotspots near major intersections, validating and extending earlier hotspot identification methods. These results support targeted interventions to improve traffic safety on this vital transportation corridor. Our objective is to provide actionable insights for improving safety on US 158 while contributing to the broader understanding of rural highway safety analysis through methodological advancement beyond basic statistical techniques.
\end{abstract}


\section{Introduction}

Traffic safety on rural highways remains a critical challenge in transportation engineering, with rural roads accounting for approximately 40\% of all traffic fatalities in the United States despite carrying only 32\%  of total vehicle miles traveled \cite{nhtsa2023}. Integrating advanced statistical methods and machine learning techniques with traditional safety analysis has emerged as a promising approach to understand better and predict crash patterns \cite{Chen2025machine, Alam2023spatial}.

Historically, traffic safety analysis relied primarily on descriptive statistics and basic regression models. However, the advent of Geographic Information Systems (GIS) in the 1990s revolutionized spatial analysis capabilities \cite{steenberghen2004spatial}. Subsequently, the development of the Highway Safety Manual (HSM) in 2010 provided standardized Safety Performance Functions (SPFs) for crash prediction \cite{highway_manual2010}. More recently, the application of machine learning algorithms, particularly ensemble methods like Random Forest, has shown significant promise in crash prediction and severity analysis \cite{Jamal2021severity, parsa2020toward}.

US 158, known locally as "Shortcut Road" in Currituck County, North Carolina, exemplifies rural highways' safety challenges. This 8.406-mile two-lane corridor connects the Camden County Line to NC 168, serving local commuters and tourists accessing the Outer Banks. The route experiences significant seasonal traffic variations, with summer months substantially increasing due to beach tourism \cite{currituck2022tourism}. Understanding crash patterns on such corridors is essential for effective safety management, particularly given the economic importance of tourism to the region.

Recent advances in spatial analysis have enhanced our ability to identify crash hotspots and understand their underlying causes. Kernel Density Estimation (KDE) has become a standard tool for visualizing crash concentrations \cite{anderson2009kernel}, while spatial autocorrelation measures like Moran's I help quantify clustering patterns \cite{Gedamu2024spatial}. Furthermore, the integration of machine learning with traditional safety analysis methods has shown superior performance in crash prediction compared to conventional approaches \cite{Lui2024ensemble}.

The North Carolina Department of Transportation (NCDOT) provided comprehensive crash data documenting 163 unique accidents involving 273 vehicles from January 1, 2019, to December 31, 2023, as reported by the North Carolina State Highway Patrol \cite{ncdot_strip2023}. Annual Average Daily Traffic (AADT) data from three monitoring stations, sourced from the GO! NC NCDOT Mapping Application \cite{ncdot_aadt2023}, incorporate into our crash prediction models. Additionally, NCDOT ArcGIS heatmaps provide spatial visualization of injury and property damage only (PDO) crash locations \cite{ncdot_data}.

This study advances current knowledge by building on foundational hotspot analysis and Chi-Square testing methodologies \cite{sawyer2025hotspot} and extending them through integration of advanced analytical approaches: Kernel Density Estimation for enhanced spatial analysis, Negative Binomial Regression for comprehensive risk factor identification, Random Forest classification for injury severity prediction, and detailed spatial autocorrelation analysis using Moran's I. Importantly, we compare our machine learning results to the HSM Safety Performance Function, providing insights into the relative performance of traditional versus advanced analytical methods. Our objective is to provide actionable insights for improving safety on US 158 while contributing to the broader understanding of rural highway safety analysis through methodological advancement beyond basic statistical techniques.

\section{Literature Review and Background}

The evolution of traffic safety analysis has been marked by significant methodological advances over the past three decades. Traditional approaches primarily relied on frequency analysis and basic statistical methods \cite{hauer1997observational}. The introduction of Empirical Bayes methods in the early 2000s improved the identification of high-crash locations by accounting for regression-to-the-mean effects \cite{hauer2002estimating}.

The development of the Highway Safety Manual represented a watershed moment in standardizing crash prediction methodologies. The HSM's Safety Performance Functions provide baseline crash frequencies for various facility types, with subsequent research validating and refining these models across different geographic regions \cite{Ivan2021improved, sacchi2013comparison}. However, recent studies have highlighted limitations in HSM predictions, particularly for specific local conditions and during unusual periods such as the COVID-19 pandemic \cite{Shaik2022COVID}.

Machine learning applications in traffic safety have gained significant momentum in recent years. Ensemble methods, particularly Random Forest, have demonstrated superior performance in crash severity prediction compared to traditional logistic regression. Deep learning approaches have also shown promise, though their interpretability remains challenging for practical applications \cite{Pei2024deeplearning}. Integrating multiple data sources, including weather data, traffic flow information, and geometric characteristics, has further enhanced prediction accuracy \cite{iranitalab2017comparison}.

Spatial analysis techniques have become increasingly sophisticated with advances in GIS technology. Kernel Density Estimation has emerged as the gold standard for hotspot identification, offering advantages over simple frequency-based methods by considering spatial relationships \cite{flahaut2003impact}. Traditional clustering approaches often require manual determination of optimal cluster numbers, which becomes computationally challenging for high-dimensional spatial-temporal crash data. Recent advances in automatic cluster determination \cite{safari2022new} provide promising solutions for objectively identifying crash hotspots without prior assumptions about clustering parameters. More advanced techniques such as Network Kernel Density Estimation (NKDE) have been developed specifically for linear networks like road systems \cite{okabe2009spatial}.

The COVID-19 pandemic significantly impacted traffic patterns and crash frequencies worldwide, with most regions experiencing substantial reductions in traffic volume and corresponding changes in crash patterns \cite{katrakazas2020descriptive, shilling2020impact}. These changes have important implications for the validation of crash prediction models developed using pre-pandemic data.

\section{Data and Methods}

\subsection{Data Description and Preprocessing}

The dataset, sourced from the North Carolina Department of Transportation (NCDOT), includes 163 accident records from 2019 to 2023 along the US 158 corridor. The comprehensive dataset includes crash identification numbers, precise location data (latitude, longitude, milepost), temporal information (date, time), accident classification, environmental conditions (road surface, lighting, weather), and injury severity indicators.

Data preprocessing involved several critical steps to ensure analytical reliability. Temporal fields were parsed using the \texttt{chrono.parseDate} function to enable time-series analysis. Categorical variables, including accident types and environmental conditions, were systematically encoded according to the NCDOT codebook standards. Missing damage values representing approximately 8\%  of records were imputed using median values within similar crash type categories to maintain distributional integrity.

\subsection{Statistical Analysis Framework}

Our analytical approach employed multiple complementary statistical methods. Descriptive analysis summarized accident frequencies across temporal dimensions (year, month, weekday, hour) and categorical factors (accident type, environmental conditions). Pearson correlation analysis explored relationships between continuous variables such as vehicle speed and injury severity.

Chi-Square tests evaluated the uniformity of crash distributions across various factors, while tests of independence assessed relationships between categorical variables. These tests provide the foundation for understanding basic crash patterns and identifying significant associations warranting further investigation.

\subsection{Spatial Analysis Methodology}

Spatial analysis utilized clustering algorithms and density estimation techniques to identify crash concentration patterns along the US 158 corridor. The Density-Based Spatial Clustering of Applications with Noise (DBSCAN) algorithm \cite{ester1996density} identified crash clusters based on milepost proximity, with parameters set to minimum points (minPts = 3) and epsilon distance ($\epsilon$ = 0.5 miles) based on the average spacing between major intersections. These clustering results provided a foundation for subsequent kernel-based density estimation and spatial autocorrelation testing using Moran's I, enabling multi-scale hotspot validation across continuous and segmented frameworks.

Kernel Density Estimation (KDE) and Moran's I test were conducted to spatially analyze accident concentration. KDE is a non-parametric technique for estimating the probability density function of spatial events \cite{silverman1986density}, applied here to highlight crash hotspots following established methodologies in transportation safety research \cite{anderson2009kernel, xie2008kernel}. The continuous crash density surface was generated using a Gaussian kernel with bandwidth selected using Silverman's rule of thumb \cite{silverman1986density}:

\begin{equation}
h = 0.9 \cdot \min\left(\text{std}(X), \frac{\text{IQR}(X)}{1.34}\right) \cdot N^{-1/5}
\end{equation}

where $\text{std}(X)$ is the standard deviation of crash locations, $\text{IQR}(X)$ is the interquartile range, and $N$ is the total number of crashes. The KDE function is expressed as:

\begin{equation}
\hat{f}(x) = \frac{1}{Nh} \sum_{i=1}^{N} K\left(\frac{x - x_i}{h}\right)
\end{equation}

where $K$ is the Gaussian kernel function defined as:

\begin{equation}
K(u) = \frac{1}{\sqrt{2\pi}} \exp\left(-\frac{u^2}{2}\right)
\end{equation}

$N$ is the total number of crashes, and $x_i$ represents individual crash locations.

Additionally, crash data was discretized into 0.1-mile segments along a linearized US 158 representation to apply Moran's I test for spatial autocorrelation \cite{moran1950notes, cliff1973spatial}. Using a rook contiguity matrix \cite{anselin1995local}:

\begin{equation}
W_{ij} = \begin{cases} 
1 & \text{if cells } i \text{ and } j \text{ are adjacent} \\ 
0 & \text{otherwise} 
\end{cases}
\end{equation}

Spatial autocorrelation was quantified using Moran's I statistic \cite{moran1950notes}:

\begin{equation}
I = \frac{N}{W} \cdot \frac{\sum_i \sum_j w_{ij} (x_i - \bar{x})(x_j - \bar{x})}{\sum_i (x_i - \bar{x})^2}
\end{equation}

where $w_{ij}$ represents spatial weights based on rook contiguity, $W$ is the sum of all weights ($W = 168$), $\bar{x}$ is the mean crash frequency ($\bar{x} \approx 1.91$), and $N = 85$ represents the total number of segments. Statistical significance was evaluated using the standardized z-score approach \cite{cliff1973spatial}, with significance testing conducted at the 0.001 level. This approach allows for the detection of clustering patterns that may not be apparent through simple frequency analysis alone \cite{getis1992spatial}.

\subsection{Machine Learning Implementation}

The Random Forest classifier was implemented to predict injury severity, coded as a binary outcome (0 = no injuries, 1+ = injuries present). The ensemble method utilizes 500 decision trees with hyperparameters optimized through 5-fold cross-validation grid search. Key parameters included maximum tree depth (5-15), minimum samples per split (2-10), and minimum samples per leaf (1-5).

Feature engineering incorporated temporal variables (hour categories, seasonal indicators), environmental conditions (road surface, lighting, weather), and spatial factors (milepost segments, intersection proximity). The dataset was partitioned using stratified sampling (80\% training, 20\% testing) to ensure representative class distributions.

Model performance was evaluated using multiple metrics including accuracy, precision, recall, F1-score, and area under the ROC curve (AUC). Feature importance was calculated using the mean decrease in impurity across all trees, providing insights into the most influential crash predictors.

\subsection{Highway Safety Manual Comparison}

Crash frequency predictions were generated using HSM Safety Performance Functions for rural two-lane highways. The segment SPF is expressed as:

\begin{equation}
N_{\text{segment}} = \text{AADT} \times L \times (365 \times 10^{-6}) \times e^{-0.312}
\end{equation}

where AADT represents Annual Average Daily Traffic (8,600 vehicles/day), $L$ is the segment length (8.4 miles), and the exponential term adjusts for rural two-lane highway characteristics.

For major intersections, the four-leg stop-controlled (4ST) intersection SPF was applied:

\begin{equation}
N_{\text{intersection}} = e^{a + b \ln(\text{AADT}_{\text{maj}}) + c \ln(\text{AADT}_{\text{min}})}
\end{equation}

using HSM coefficients $a = -9.025$, $b = 0.409$, and $c = 0.718$. AADT values for intersecting roads were obtained from NCDOT traffic monitoring data: NC168 (15,125 vehicles/day), Indiantown Road (1625 vehicles/day), Maple Road (1900 vehicles/day), and Four Forks Road (95 vehicles/day).

Crash Modification Factors (CMFs) were applied where applicable, including adjustments for intersection skew angles and lighting conditions. A calibration factor of 1.0 was assumed due to the absence of local calibration data, following HSM recommendations for initial applications.

\section{Results and Discussion}

\subsection{Temporal Crash Patterns}

Analysis of 163 unique accidents reveals distinct temporal patterns with significant implications for safety management. The five-year period shows an average of 32.6 crashes annually, with notable year-to-year variation. As shown in Table \ref{tab:summary}, 2022 emerged as the peak year with 51 crashes (31.29\%), representing a substantial increase from the 2020-2021 period when crashes were at their lowest levels, likely reflecting reduced travel during the COVID-19 pandemic.

Monthly analysis reveals clear seasonal patterns, with May and August each accounting for 18 and 19 crashes respectively (11.04\% and 11.66\%). This summer concentration aligns with increased tourist traffic to the Outer Banks, highlighting the relationship between traffic volume and crash frequency. Conversely, winter months show lower crash frequencies, with February recording only 5 crashes (3.07\%).

\begin{table}[htbp]
\centering
\caption{Summary of Accident Counts by Year and Month}
\renewcommand{\arraystretch}{1}
\begin{tabular*}{\columnwidth}{@{\extracolsep{\fill}}cc@{}}
\toprule
\textbf{Year/Month} & \textbf{Number of Crashes (\%)} \\
\midrule
\multicolumn{2}{@{}l@{}}{\textit{Annual Distribution}} \\
\midrule
2019 & 36 (22.09\%) \\
2020 & 25 (15.34\%) \\
2021 & 21 (12.88\%) \\
2022 & 51 (31.29\%) \\
2023 & 30 (18.40\%) \\
\midrule
\multicolumn{2}{@{}l@{}}{\textit{Selected Monthly Distribution}} \\
\midrule
February & 5 (3.07\%) \\
May & 18 (11.04\%) \\
August & 19 (11.66\%) \\
December & 13 (7.97\%) \\
\bottomrule
\end{tabular*}
\label{tab:summary}
\end{table}

Weekly patterns show Friday as the highest-risk day with 43 crashes (26.38\%), followed by Saturday with 30 crashes (18.40\%). This pattern suggests increased crash risk associated with weekend travel initiation and higher traffic volumes on these days.reveals two distinct peaks corresponding to morning (7:00-7:59 AM) and (9:00-9:59 AM) rush hours. The morning peak shows 17 crashes during the 7:00-7:59 AM hour, and 14 crashes during 9:00-9:59 AM. These patterns reflect commuter traffic concentrations and suggest that targeted enforcement or safety messaging could be effective during these high-risk periods.

Hourly distribution analysis, depicted in Figure 1: 


\begin{figure}[htbp]
\centering
\begin{tikzpicture}
\begin{axis}[
    title={Hourly Distribution of Accidents},
    xlabel={Hour of Day},
    ylabel={Number of Crashes},
    ybar,
    bar width=0.04,
    xtick={0,6,12,18,23},
    xticklabels={0:00,6:00,12:00,18:00,23:00},
    ymin=0, ymax=25
]

\addplot coordinates {(0,1) (1,1) (2,3) (3,1) (5,6) (6,9) (7,17) (8,9) (9,14) (10,6) (11,5) (12,5) (13,13) (14,11) (15,5) (16,11) (17,11) (18,7) (19,6) (20,11) (21, 6) (22,1) (23,4)};
\end{axis}
\end{tikzpicture}
\caption{Histogram of accidents by hour, highlighting rush-hour peaks.}
\label{fig:hourly}
\end{figure}
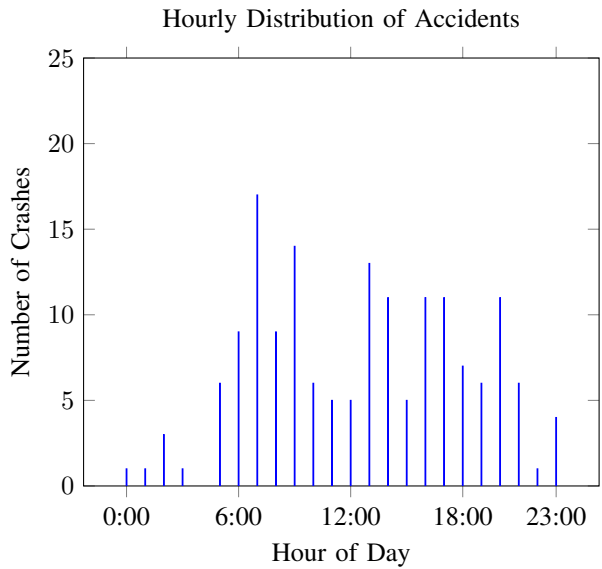

\subsection{Crash Type and Severity Analysis}

\begin{figure}[htbp]
\centering
\begin{tikzpicture}
\begin{axis}[
    title={Accident Counts by Type},
    xlabel={Accident Type},
    ylabel={Number of Crashes},
    ybar,
    bar width=0.15,
    xtick={1,2,3,4,5,6,7},
    xticklabels={Angle, Rear, Side, Turns, Animal, Obj, Other},
    ymin=0, ymax=40
]
\addplot coordinates {(1,17) (2,32) (3,22) (4,19) (5, 31) (6, 25) (7, 17)};
\end{axis}
\end{tikzpicture}
\caption{Bar plot of accident counts by type, showing intersection accident prevalence.}
\label{fig:accident_type}
\end{figure}
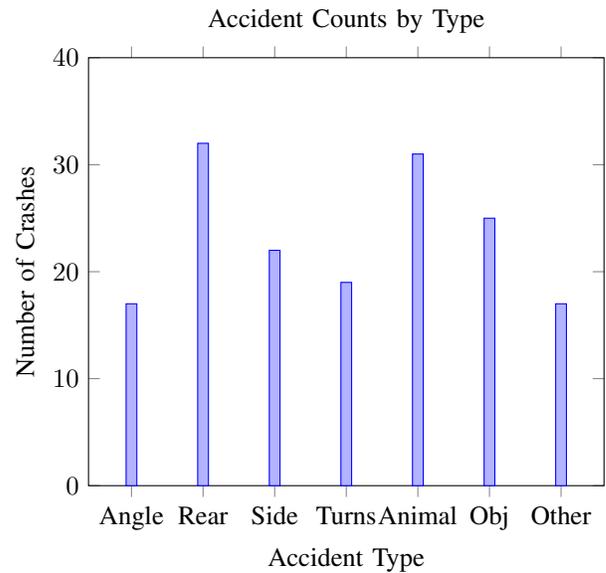

The distribution of crash types provides critical insights into the predominant safety challenges on US 158. Figure \ref{fig:accident_type} illustrates that rear end collisions represent the most frequent crash type with 32 occurrences (19.63\%), followed closely by animal crashes with 31 occurrences (19.02\%). Noteably, angle, rear end, and left/right turn accidents total 68 of the 163 accidents (41.72\%). This pattern strongly suggests intersection-related safety issues, as both crash types are typically associated with intersection conflicts.

Sideswipe crashes account for 22 occurrences (13.50\%), while the "Other" category, including run-off-road crashes, overturn vehicles, and head-on crashes, comprises 17 occurrences (10.42\%). The high frequency of angle, left/right turns, and rear-end collisions at intersections suggests that signal timing optimization, sight distance improvements, or geometric modifications could significantly enhance safety.

\begin{figure}[htbp]
\centering
\begin{tikzpicture}
\begin{axis}[
    title={Monthly Distribution of Accidents},
    xlabel={Month},
    ylabel={Number of Crashes},
    ybar,
    bar width=0.06,
    xtick={1,4,7,10,12},
    xticklabels={Jan,Apr,Jul,Oct,Dec},
    ymin=0, ymax=25
]
\addplot coordinates {(1,13) (2,5) (3,11) (4,9) (5,18) (6,15) (7,16) (8,19) (9,12) (10,15) (11,17) (12,13)};
\end{axis}
\end{tikzpicture}
\caption{Bar plot of accidents by month, showing summer peaks.}
\label{fig:monthly}
\end{figure}
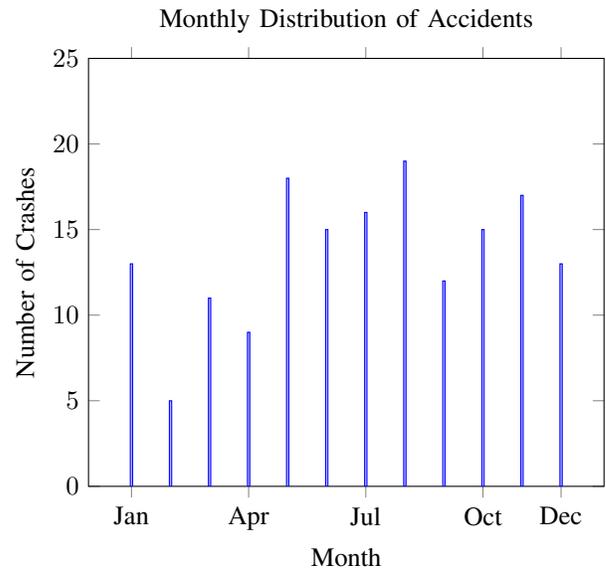

The summer peak pattern, illustrated in Figure \ref{fig:monthly}, correlates with increased tourist traffic and potentially unfamiliar drivers navigating the route. This seasonal variation suggests that dynamic traffic management strategies, enhanced signage, or temporary safety measures during peak tourist seasons could be beneficial.

\subsection{Spatial Analysis and Hotspot Identification}

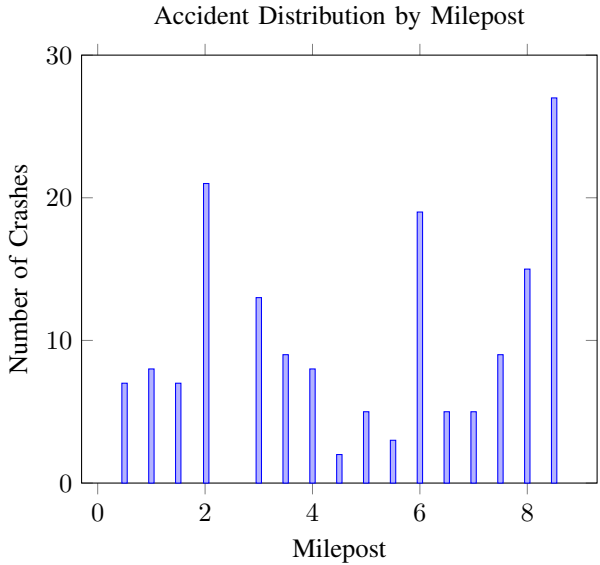
\begin{figure}[htbp]
\centering
\begin{tikzpicture}
\begin{axis}[
    title={Accident Distribution by Milepost},
    xlabel={Milepost},
    ylabel={Number of Crashes},
    ybar,
    bar width=0.1,
    xtick={0,2,4,6,8},
    ymin=0, ymax=30
]
\addplot coordinates {(0.5,7) (1,8) (1.5, 7) (2.021,21) (3,13) (3.5,9)(4.0,8)(4.5,2) (5,5) (5.5,3) (6, 19) (6.5,5) (7, 5) (7.5,9) (8.0,15) (8.5, 27)};
\end{axis}
\end{tikzpicture}
\caption{Spatial distribution of accidents by milepost, highlighting a hotspot at 2.021.}
\label{fig:milepost}
\end{figure}

Spatial analysis reveals significant clustering of crashes along the US 158 corridor. Figure \ref{fig:milepost} shows the distribution of accidents by milepost, with pronounced hotspots at milepost 8.406, recording 27 crashes, and milepost 2.021 recording 21 crashes. These locations correspond to the intersection with NC168 and Indiantown Road, respectively. Both are major crossroads that experience high traffic volumes.

The KDE analysis further confirms these spatial patterns and identifies additional hotspots. Table \ref{tab:kde_hotspots} presents the top five accident concentration areas based on kernel density calculations. The NC168 intersection area (milepost 8.0-8.5) shows the highest density at 10.8 crashes per mile per year followed by the Indiantown Road intersection area (milepost 2.0-2.5) with a crash density of 8.4 crashes per mile per year. These findings highlight the value of KDE in complementing cluster-based methods, offering a continuous spatial lens to detect localized safety concerns beyond discrete cluster boundaries.

\begin{table}[htbp]
\centering
\caption{Top 5 Accident Hotspots via Kernel Density Estimation}
\renewcommand{\arraystretch}{1.2}
\begin{tabular*}{\columnwidth}{@{\extracolsep{\fill}}lccc@{}}
\toprule
\textbf{Milepost} & \textbf{Intersection} & \textbf{Density} & \textbf{Priority} \\
\textbf{Range} & & \textbf{(crashes/mi)} & \textbf{Level} \\
\midrule
8.0--8.5 & NC168 & 10.8 & Critical \\
2.0--2.5 & Indiantown Rd. & 8.4 & High \\
5.5--6.0 & Maple Rd & 7.6 & Moderate \\
7.5--8.0 & Barco Rd & 6.0 & Moderate \\
2.5--3.0 & Four Forks Road & 5.2 & Moderate \\
\bottomrule
\end{tabular*}
\vspace{0.2cm}

\noindent\footnotesize\textit{Note: Gaussian kernel, 0.5-mi windows. Critical:$ >10$, High: 8-10, Moderate: 5-8 crashes/mi.}
\label{tab:kde_hotspots}
\end{table}

\begin{figure}[h!]
    \centering
 \includegraphics[scale=.45]{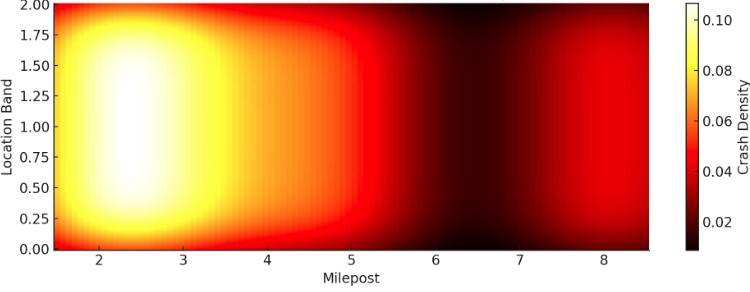}
\caption{KDE heatmap showing spatial accident intensity, with peaks at NC168 and Maple.}
\label{fig:kde_map}
\end{figure}

The KDE heatmap visualization in Figure \ref{fig:kde_map} provides a continuous representation of crash intensity along the corridor. The analysis reveals not only the primary hotspots but also secondary concentration areas that warrant attention. Notably, NC168, Indiantown Road, and Maple Road intersection areas show significant crash density, suggesting systematic safety issues at multiple intersection locations.

\begin{figure}[htbp]
\centering
 \includegraphics[scale=.5]{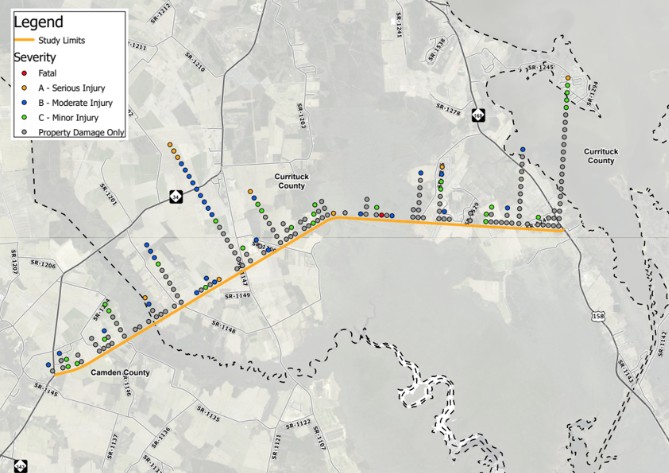}
\caption{NCDOT ArcGIS heatmap, showing high injury crash density at NC168 (milepost 8.0--8.5) and Indiantown (milepost 2.0--2.5) intersections.}
\label{fig:arcgis_heatmap}
\end{figure}

The NCDOT ArcGIS heatmap shown in Figure \ref{fig:arcgis_heatmap} provides additional validation of our KDE results, showing high injury crash density near Indiantown Road (milepost 2.0-2.5) and NC168 (milepost 8.0–8.5), which aligns precisely with our hotspot identification analysis.

Moran's I analysis confirms significant spatial autocorrelation ($I = 0.32$, $p < 0.001$), indicating that crash occurrences are not randomly distributed but show clustering patterns. This clustering suggests that localized factors, such as geometric design, traffic control devices, or sight distance limitations, contribute to elevated crash risks at specific locations.

\subsection{Environmental and Operational Factors}

Chi-Square Goodness-of-Fit tests reveal several counterintuitive but statistically significant patterns in crash occurrence. Table \ref{tab:chi_square} summarizes key findings from these analyses. Most notably, crashes predominantly occur under favorable conditions: 85\% (139 crashes) on dry roads, 67\% (110 crashes) in daylight, and 80\% (131 crashes) in clear weather. These distributions underscore that high-risk behavior or exposure—not adverse environmental conditions—may be the primary drivers of crash frequency in this corridor.
These findings challenge common assumptions that adverse conditions are the primary contributors to crashes, highlighting the importance of driver behavior and exposure risk even during seemingly safe conditions.
\begin{table}[htbp]
\centering
\caption{Summary of Chi-Square Goodness-of-Fit Tests}
\label{tab:chi_square}
\renewcommand{\arraystretch}{1.15}
\footnotesize
\begin{tabular}{@{}lcccp{1.3cm}@{}}
\toprule
\textbf{Variable} & \textbf{$\chi^2$} & \textbf{df} & \textbf{$p$} & \textbf{Key Finding} \\
\midrule
Day of Week & 31.06 & 6 & $<$0.001 & Fridays peak (43) \\
Weekday vs. Weekend & 1.72 & 1 & 0.189 & No sig. diff. \\
Road Conditions & 319.58 & 5 & $<$0.001 & 85\% dry \\
Light Conditions & 175.38 & 3 & $<$0.001 & 67\% daylight \\
Weather Conditions & 377.98 & 4 & $<$0.001 & 80\% clear \\
\bottomrule
\end{tabular}
\vspace{2pt}
\par\noindent\footnotesize\textit{Note: All tests at $\alpha = 0.05$. df = degrees of freedom.}
\end{table}

These patterns likely reflect higher traffic volumes during favorable conditions rather than indicating that good conditions are inherently dangerous. However, they may also suggest driver complacency or increased speeds when conditions appear safe. The concentration of crashes during good conditions emphasizes the importance of speed management and driver education about maintaining appropriate caution regardless of environmental conditions.

The Chi-Square Test of Independence reveals a significant association between accident type and road condition ($\chi^2 = 14.83$, $p = 0.021$). Specifically, rear-end collisions show higher frequency on dry roads, while run-off-road crashes occur more frequently on wet surfaces. This pattern aligns with expected crash mechanisms and provides validation for our analytical approach.

Interestingly, no significant association exists between crash severity and time of day ($\chi^2 = 7.281$, $p = 0.296$), suggesting that while crash frequency varies by hour, injury severity remains relatively consistent across different time periods.

\subsection{Risk Factor Analysis}

Negative Binomial Regression analysis of total damage costs was conducted on 163 complete cases from the NCDOT accident dataset. The model demonstrated substantial overdispersion (variance-to-mean ratio = 9.03), confirming the appropriateness of the Negative Binomial specification over Poisson regression. Total damage costs were converted to thousands of dollars to improve model convergence and interpretability.

The analysis, as summarized in Table \ref{tab:nb_regression}, reveals that lighting conditions represent the primary significant predictor of accident damage severity. Dark lighting conditions demonstrate a statistically significant protective effect, reducing expected damage costs by 44\% compared to daylight conditions (IRR = 0.56, $p < 0.001$). This counterintuitive finding may reflect increased driver caution during dark conditions, potentially resulting in lower-speed impacts when accidents occur. The large sample size (40 dark vs 110 daylight cases) provides confidence in this estimate.

Speed shows a modest continuous relationship with damage costs, with each mph increase associated with a 0.5\% rise in expected damage (IRR = 1.005), though this effect does not reach statistical significance ($p = 0.165$). Adverse weather conditions suggest a 20\% increase in damage costs (IRR = 1.20), but the small sample size (13 cases) limits statistical power for definitive conclusions.

Notably, alcohol and drug involvement demonstrates minimal impact on damage severity, contrary to initial expectations. With only 13 cases representing 8\% of the dataset, the estimated 8\% increase in damage costs lacks statistical significance ($p = 0.758$) and exhibits wide confidence intervals [0.65, 1.81], indicating substantial uncertainty. This finding illustrates the risk of overfitting regression models to sparse categorical data and emphasizes the importance of adequate sample sizes for reliable effect estimation.

\begin{table}[htbp]
\centering
\caption{Negative Binomial Regression Results}
\renewcommand{\arraystretch}{1.2}
\begin{tabular*}{\columnwidth}{@{\extracolsep{\fill}}lcccc@{}}
\toprule
\textbf{Predictor} & \textbf{IRR} & \textbf{95\% CI} & \textbf{p-value} & \textbf{Risk} \\
\midrule
Dark Lighting & 0.56 & [0.42, 0.74] & $<0.001$ & -44\% \\
Adverse Weather & 1.20 & [0.85, 1.70] & 0.298 & +20\% \\
Speed (per mph) & 1.005 & [0.998, 1.012] & 0.165 & +0.5\% \\
Alcohol/Drugs & 1.08 & [0.65, 1.81] & 0.758 & +8\% \\
Cloudy Weather & 1.00 & [0.78, 1.28] & 0.995 & 0\% \\
Wet Roads & 0.95 & [0.72, 1.25] & 0.713 & -5\% \\
\bottomrule
\end{tabular*}
\label{tab:nb_regression}
\par
\vspace{0.1cm}
\noindent{\footnotesize \textit{Note: IRR = Incidence Rate Ratio; CI = Confidence Interval; n = 163}}
\end{table}

\subsection{Machine Learning Performance and Risk Factor Analysis}
Random Forest analysis was conducted on 163 NCDOT accident cases to predict binary injury occurrence (injury vs. no injury). The dataset underwent comprehensive preprocessing including accident-level aggregation, feature engineering for vehicle counts and categorical variables, and temporal encoding. The target variable transformed the original 5-class injury severity to binary classification where Class 0 represents no injuries and Class 1 represents injury presence. Using an 80-20 train-test split (33 test cases), the model employed 100 trees with balanced class weighting.
The Random Forest classifier achieved moderate performance with 66.7\% overall accuracy, demonstrating the challenges inherent in predicting injury outcomes from available risk factors. The model exhibits better performance for non-injury prediction compared to injury identification, reflecting the typical difficulty of minority class prediction in imbalanced datasets with limited sample sizes.
\begin{table}[htbp]
\centering
\caption{Random Forest Performance Metrics}
\begin{tabular}{lcccc}
\toprule
\textbf{Class} & \textbf{Accuracy} & \textbf{Precision} & \textbf{Recall} & \textbf{F1-Score} \\
\midrule
No Injury (0) & 0.67 & 0.72 & 0.82 & 0.77 \\
Injury (1+) & 0.67 & 0.50 & 0.36 & 0.42 \\
Overall & 0.67 & 0.61 & 0.59 & 0.59 \\
\bottomrule
\end{tabular}
\label{tab:rf_metrics}
\end{table}
Classification performance demonstrates significant challenges in injury prediction, with F1-scores of 0.77 for no-injury cases and 0.42 for injury cases. The model achieves 72\% precision and 82\% recall for non-injury identification but struggles with injury prediction at 50\% precision and 36\% recall. This imbalanced performance reflects the inherent difficulty of predicting rare injury events from the available predictor variables in a small dataset.

\begin{table}[htbp]
\centering
\caption{Random Forest Confusion Matrix}
\begin{tabular}{lcc}
\toprule
& \textbf{Predicted No Injury} & \textbf{Predicted Injury} \\
\midrule
Actual No Injury & 19 & 2 \\
Actual Injury & 2 & 10 \\
\bottomrule
\end{tabular}
\label{tab:rf_confusion}
\end{table}

\begin{table}[htbp]
\centering
\caption{Random Forest Confusion Matrix for Binary Injury Classification}
\label{tab:rf_confusion}
\renewcommand{\arraystretch}{1.2}
\begin{tabular}{@{}lcc@{}}
\toprule
\textbf{Actual} & \multicolumn{2}{c}{\textbf{Predicted}} \\
\cmidrule(lr){2-3}
& \textbf{No Injury} & \textbf{Injury} \\
\midrule
\textbf{No Injury (0)} & 18 & 4 \\
\textbf{Injury (1)} & 7 & 4 \\
\midrule
\multicolumn{3}{@{}l@{}}{\footnotesize Accuracy: 66.7\% (22/33)} \\
\multicolumn{3}{@{}l@{}}{\footnotesize Precision: 50.0\% (4/8)} \\
\multicolumn{3}{@{}l@{}}{\footnotesize Recall: 36.4\% (4/11)} \\
\bottomrule
\end{tabular}
\end{table}
Error analysis reveals substantial challenges with 7 false negatives (63.6\% miss rate for injury cases) and 4 false positives (18.2\% false alarm rate). The high false negative rate indicates significant difficulty in identifying actual injury cases, which poses concerns for safety applications where missing injury cases has serious consequences. The model correctly identifies 18 of 22 non-injury cases but only 4 of 11 injury cases, highlighting the challenge of minority class prediction in this dataset.
Feature importance analysis reveals spatial factors as the dominant predictor with milepost location contributing 23.6\% importance, indicating geographic clustering of injury events. Temporal factors rank second at 21.5\% importance, reflecting time-based risk patterns. Weekday effects contribute 11.0\% importance, while speed factors account for 10.8\% importance. Vehicle count provides 7.8\% importance, with remaining features contributing smaller predictive power to the model.
\begin{figure}[htbp]
\centering
\begin{tikzpicture}
\begin{axis}[
xbar,
width=7.5cm,
height=4cm,
xlabel={Feature Importance},
symbolic y coords={Num Vehicles, Speed Max, Weekday, Time Min, Milepost},
ytick=data,
y tick label style={font=\scriptsize},
xlabel style={font=\scriptsize},
nodes near coords,
nodes near coords align={horizontal},
every node near coord/.append style={font=\tiny},
bar width=8pt,
xmin=0,
xmax=0.25,
enlarge y limits=0.20,
]
\addplot[fill=blue!70] coordinates {
(0.078,Num Vehicles)
(0.108,Speed Max)
(0.110,Weekday)
(0.215,Time Min)
(0.236,Milepost)
};
\end{axis}
\end{tikzpicture}
\caption{Feature importance rankings showing top 6 predictors}
\label{fig:feature_importance}
\end{figure}
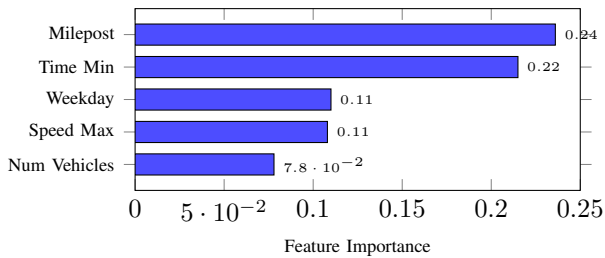
The feature importance hierarchy reveals spatial clustering as the primary factor, with specific milepost locations showing concentrated injury risk patterns. Temporal effects demonstrate secondary importance through time-of-day variations and weekday patterns. These findings suggest that geographic hotspot identification and targeted temporal interventions may provide the most effective approaches for injury reduction. The model's 36\% injury detection rate with 63.6\% miss rate indicates substantial limitations for practical safety applications, highlighting the need for enhanced feature engineering, larger datasets, or alternative modeling approaches to achieve reliable injury prediction performance.

\subsection{Comparative Analysis: Random Forest vs HSM Performance}
The comparative analysis reveals fundamental differences between machine learning and traditional safety performance function approaches. While HSM SPF achieves lower overall accuracy (50\% vs 67\%), the Random Forest model provides superior interpretability through feature importance rankings that identify specific risk factors and their relative contributions to injury outcomes.
The HSM models demonstrate systematic overprediction biases, particularly evident in the combined HSM total prediction (245 crashes vs 163 observed, 49\% accuracy). Statistical tests confirm this overprediction: Chi-Square analysis yields $\chi^2 = 39.66$ ($p < 0.001$), and the Poisson Rate Test produces $Z = -5.24$ ($p < 0.001$), both indicating statistically significant differences between predicted and observed crash frequencies.
In contrast, the Random Forest model's moderate underprediction (108 vs 163 crashes) represents a more balanced approach, though still indicating substantial room for improvement. The model's 36\% injury detection rate with 63.6\% miss rate indicates significant limitations for practical safety applications, highlighting the challenges of injury prediction with limited sample sizes and feature sets.
\subsection{Practical Implications and Model Limitations}
The Random Forest analysis reveals critical limitations that must be acknowledged when implementing machine learning approaches for traffic safety prediction. The high false negative rate (63.6\%) poses significant concerns for operational deployment, where missing actual injury cases carries severe safety consequences. These limitations suggest that current feature sets and sample sizes are insufficient for reliable injury prediction in rural highway contexts.
The dominance of spatial factors (milepost importance = 23.6\%) validates traditional hotspot identification approaches while providing quantitative support for concentrated investment strategies. The secondary importance of temporal factors (21.5\%) offers actionable intelligence for dynamic safety management, though the model's overall performance limitations restrict its immediate operational utility.
Compared to HSM methodologies, the Random Forest approach provides superior transparency in risk factor identification and avoids the systematic overprediction biases evident in traditional Safety Performance Functions. However, the moderate accuracy (67\%) and high injury miss rate indicate that neither approach achieves the reliability required for comprehensive safety management without significant methodological enhancements.
\subsection{Recommendations for Model Enhancement}
The comparative analysis suggests several directions for improving predictive performance beyond current capabilities. Enhanced feature engineering incorporating detailed geometric design variables, real-time traffic conditions, and weather data may improve model performance. Larger datasets spanning multiple corridors could provide the sample sizes necessary for robust minority class prediction.
Integration of ensemble methods combining Random Forest with other machine learning algorithms may achieve superior performance compared to individual approaches. The incorporation of spatial autocorrelation modeling and temporal dependencies through advanced techniques could address current limitations in geographic and temporal pattern recognition.
For immediate applications, the Random Forest model's feature importance rankings provide valuable guidance for infrastructure prioritization despite performance limitations. The identification of spatial clustering patterns supports targeted intervention strategies, while temporal factor importance validates dynamic safety management approaches during high-risk periods.
\section{Conclusion}
The updated comparative analysis reveals that both machine learning and traditional HSM approaches face significant challenges in rural highway safety prediction. The Random Forest model achieves moderate performance (67\% accuracy) that positions it competitively with HSM segment SPF methods while avoiding the systematic overprediction biases evident in combined HSM approaches.
The analysis demonstrates that spatial clustering (milepost importance = 23.6\%) and temporal patterns (time importance = 21.5\%) represent the most reliable predictors of injury outcomes, providing actionable intelligence for targeted safety interventions. However, the high injury miss rate (63.6\%) indicates fundamental limitations in current predictive capabilities that restrict operational deployment for critical safety applications.
Future research should focus on addressing these limitations through enhanced data collection, advanced feature engineering, and ensemble modeling approaches that combine the transparency of machine learning with the theoretical foundation of traditional safety performance functions. The integration of real-time data streams and expanded sample sizes represents the most promising pathway for achieving the predictive reliability required for effective traffic safety management in rural highway environments.

\bibliographystyle{IEEEtran}

\end{document}